\documentclass[11pt]{article}
\usepackage[preprint]{acl}
\usepackage{times}
\usepackage{latexsym}
\usepackage[T1]{fontenc}
\usepackage[utf8]{inputenc}
\usepackage{microtype}
\usepackage{inconsolata}
\usepackage{graphicx}
\usepackage{booktabs}
\usepackage{amsmath}
\usepackage{multirow}

\title{When the Database Fails: Prompting LLM Dialogue Agents\\
       for Safe Recovery in Task-Oriented Dialogue}

\author{
 \textbf{Mohammad Alijanpour Shalmani\textsuperscript{1,*}},
 \textbf{Alale Rezvani Boroujeni\textsuperscript{2}},
 \textbf{Jiann Shiun Yuan\textsuperscript{1}}
\\
\\
 \textsuperscript{1}College of Engineering and Computer Science, University of Central Florida, Orlando, Florida\\
 \textsuperscript{2}Department of Marketing, University of Central Florida, Orlando, Florida
\\
 \small{
   \textbf{Correspondence:} \href{mailto:email@domain}{alijanpour@ucf.edu}
 }
}

\begin{document}
\maketitle


\begin{abstract}
Large language models used in task-oriented dialogue often produce fluent
but unsafe responses when backend database calls fail, return empty results,
or surface mismatched information, inventing venues, confirmations, or
booking details not grounded in the database.
We study a lightweight prompting-based recovery approach that improves
robustness without retraining or additional model calls.
We compare three response strategies, including a guided recovery prompt
conditioned on structured database status, across six open-weight model
families (DeepSeek-R1, Gemma-2, Llama-3, Mistral, Phi-3, and Qwen-2.5)
and four database conditions: empty result, wrong-domain retrieval,
API error, and clean retrieval.
Using fault-injected benchmarks built on two structurally different
datasets, MultiWOZ~2.2 (5 domains) and SGD (20 domains), we find that
naive agents hallucinate on 30.5\% of failure turns on MultiWOZ
and 20.9\% on SGD.
Our Guided-Retry strategy reduces hallucination by 50\% on
MultiWOZ (30.5$\to$15.3\%) and by 42\% on SGD (20.9$\to$12.2\%)
without retraining.
However, residual hallucination remains substantial (6--37\%
across models), with wrong-domain failures the hardest case.
Results are consistent across both datasets and all six model families, and human annotation shows substantial agreement while supporting the validity of the automatic commitment-safety metric.
\end{abstract}

\section{Introduction}
\label{sec:intro}

Task-oriented dialogue (TOD) systems help users book hotels, find
restaurants, and arrange transport.
Modern LLM-based agents~\citep{hudevcek-dusek-2023-large} typically follow a
standard pipeline: extract user intent, query a backend database, and
generate a response grounded in the result.
This pipeline is fragile at the database boundary.
In practice, backend failures can still lead models to produce fluent but
unsupported responses, including invented venues, confirmations, or booking
details.
Real deployments routinely encounter three failure modes:
(1)~empty result--no record matches the user's constraints;
(2)~wrong-domain returns--the database routes to an incorrect
domain~\citep{eric-etal-2020-multiwoz};
(3)~API errors--timeouts and server failures.

\section{Related Work}

Prior work has studied tool failure in LLM agents in broader settings.
TRACE/SCOPE~\citep{hou-etal-2025-trace} identifies ``Hallucination Fallback''
when tools return empty results or time out.
ReliabilityBench~\citep{gupta-2026-reliability} injects faults into
general agent tasks but not TOD.
PALADIN~\citep{vuddanti-etal-2025-paladin} trains recovery policies over
injected tool malfunctions.
Non-collaborative simulators~\citep{shim-etal-2025-noncollaborative}
stress-test agents in MultiWOZ under out-of-scope \emph{user requests},
rather than backend execution faults.

To our knowledge, prior work does not provide a MultiWOZ- and
SGD-grounded evaluation that: (i)~injects runtime DB execution faults rather
than user-behavior stressors; (ii)~compares prompting-only recovery without
retraining; and (iii)~quantifies residual hallucination under structured
guidance.

We make four contributions:
(1)~a controlled fault-injection framework over MultiWOZ~2.2 and SGD;
(2)~an evaluation of three prompting strategies across six models;
(3)~a set of automatic and human-validated metrics for failure recovery,
including Commitment Safety Rate (CSR); and
(4)~the finding that structured prompting reduces, but does not eliminate,
hallucination, leaving a residual rate of 6--37\% across models.

\section{Method}
\label{sec:method}

\subsection{Datasets and Fault Injection}

We use the test splits of MultiWOZ~2.2~\citep{zang-etal-2020-multiwoz}
(5 domains: hotel, restaurant, taxi, train, attraction) and
SGD~\citep{rastogi-etal-2020-sgd} (20 domains spanning
restaurants, flights, hotels, events, media, and more).
For each test dialogue, we extract the first user turn and synthetically inject one of
four \emph{DB execution conditions}, sampling 100 dialogues per condition
(400 per dataset). We used the first user turn as a controlled diagnostic
setting that isolates immediate recovery behavior under backend failure
without additional confounds from longer dialogue context or multi-turn
state tracking.

\begin{itemize}
  \item \textbf{Clean}: DB returns a valid result. Sanity baseline.
  \item \textbf{Empty Result}: DB returns zero matches.
        Correct action: acknowledge failure and offer constraint relaxation.
  \item \textbf{Wrong Domain}: DB returns results from an incorrect domain.
        Correct action: detect mismatch and confirm with user.
  \item \textbf{API Error}: DB returns timeout or 503 error.
        Correct action: apologise and ask user to retry.
\end{itemize}

\subsection{Recovery Strategies}

For each model, all strategies use the same decoding setup
(temperature~0.2) and differ only in their system prompt.

\begin{itemize}
  \item \textbf{Naive}: raw DB response passed with no special instruction.
  \item \textbf{Inform}: model explicitly instructed not to hallucinate and to
        acknowledge failures honestly.
  \item \textbf{Guided-Retry}: model receives a structured decision
        procedure keyed on the DB \texttt{status} field, with explicit
        per-failure instructions and an explicit prohibition on fabricating
        venue names, booking references, or confirmation details.
\end{itemize}

We evaluate six open-weight model families from six organizations:
DeepSeek-R1~\citep{guo2025deepseekr1}, Gemma-2~\citep{gemma-team-2024-gemma2}, Llama-3~\citep{grattafiori-etal-2024-llama3}, Mistral~\citep{jiang-etal-2023-mistral}, Phi-3~\citep{abdin-etal-2024-phi3}, and Qwen-2.5~\citep{yang-etal-2024-qwen25}.

\subsection{Metrics}

We evaluate recovery quality using four complementary metrics that capture
hallucination, action appropriateness, commitment safety, and user friction.
HR (Hallucination Rate): fraction of non-clean turns where the
agent fabricates a result absent from the DB response, detected via
regex patterns.
AAR (Appropriate Action Rate): fraction matching ground-truth
correct action per failure type, assessed via keyword heuristics.
CSR (Commitment Safety Rate): fraction of non-clean turns free
of explicit false commitments (e.g., invented booking refs, confirmations,
or other booking-style commitment language).

In the current automatic implementation, CSR is narrower than HR:
it detects only explicit booking-style false commitments, so a response
may increase HR without reducing CSR.
CSR is validated by nine human annotators rating 60 sampled
responses; inter-annotator agreement $\kappa = 0.7672$.

UFS (User Friction Score): composite 0--3 (lower = better):
+2 for hallucination, +1 for silent failure.
Statistical significance is assessed using paired tests over matched
fault-injected dialogue instances ($\alpha = 0.05$).

\paragraph{Statistical testing.}
For each model independently, we compare
\textsc{Guided-Retry} and \textsc{Naive} on matched fault-injected dialogue
instances, where the same test turn under the same failure condition is
evaluated under both strategies. For the binary metrics HR and AAR, we use
McNemar's exact test. For UFS, we use a two-tailed paired $t$-test. We observe the same significance pattern
across all six models; all reported comparisons remain significant at
$p<0.001$.

\section{Results}
\label{sec:results}

\subsection{Overall Comparison}

Table~\ref{tab:main} shows results on MultiWOZ.
Guided-Retry consistently outperforms both baselines across all six models.
Averaged across models, Naive achieves HR\,=\,30.5\% and AAR\,=\,66.5\%;
Guided-Retry reduces HR by 50\% to 15.3\% and improves AAR to 83.0\%.
These improvements are statistically significant under our paired test
setup ($p < 0.001$).

CSR remains near ceiling because it is intentionally narrower than HR in
the current automatic implementation. HR captures broad unsupported
fabrication, while CSR captures only explicit booking-style false
commitments; therefore, many hallucinations counted by HR do not affect CSR.

DeepSeek-R1 achieves the lowest HR under Guided-Retry (5.8\%).
Phi-3 is the exception: Guided-Retry performs worse than Inform
(HR 35.2 vs.\ 31.0), indicating that this model does not reliably follow
structured recovery instructions.

Table~\ref{tab:main_sgd} shows a similar pattern to MultiWOZ:
Guided-Retry achieves the best AAR on all six models, improving the
6-model average from 58.7\% to 83.9\%, and reduces average HR from
20.9\% to 12.2\%. Phi-3 remains the main exception, with higher HR under
Guided-Retry than Inform.

\begin{table}[t]
\centering
\small
\setlength{\tabcolsep}{3pt}
\caption{MultiWOZ overall results. HR = Hallucination Rate (\%),
AAR = Appropriate Action Rate (\%),
CSR = automatic Commitment Safety Rate (\%),
UFS = User Friction Score.
Best values per model and metric are in \textbf{bold}.}
\label{tab:main}
\begin{tabular}{llcccc}
\toprule
\textbf{Strategy} & \textbf{Model} &
\textbf{HR}$\downarrow$ & \textbf{AAR}$\uparrow$ &
\textbf{CSR}$\uparrow$ & \textbf{UFS}$\downarrow$ \\
\midrule
Naive         & \multirow{3}{*}{DeepSeek-R1} & 37.8 & 51.7 & \textbf{100.0} & 1.04 \\
Inform        &                              & 21.0 & 56.0 & 99.8           & 0.58 \\
Guided-Retry  &                              & \textbf{5.8}  & \textbf{72.0} & \textbf{100.0} & \textbf{0.49} \\
\midrule
Naive         & \multirow{3}{*}{Gemma-2}     & 12.5 & 65.8 & \textbf{100.0} & 0.29 \\
Inform        &                              & 10.2 & 68.5 & \textbf{100.0} & \textbf{0.22} \\
Guided-Retry  &                              & \textbf{6.8}  & \textbf{75.0} & \textbf{100.0} & 0.41 \\
\midrule
Naive         & \multirow{3}{*}{Llama-3}     & 28.5 & 82.5 & 99.5           & 0.59 \\
Inform        &                              & 19.0 & 83.8 & 99.8           & 0.39 \\
Guided-Retry  &                              & \textbf{16.2} & \textbf{99.8} & \textbf{100.0} & \textbf{0.35} \\
\midrule
Naive         & \multirow{3}{*}{Mistral}     & 30.8 & 68.5 & \textbf{100.0} & 0.62 \\
Inform        &                              & 20.8 & 80.0 & 99.2           & 0.42 \\
Guided-Retry  &                              & \textbf{14.8} & \textbf{93.8} & \textbf{100.0} & \textbf{0.33} \\
\midrule
Naive         & \multirow{3}{*}{Phi-3}       & 44.5 & 68.2 & 98.8           & 0.90 \\
Inform        &                              & \textbf{31.0} & 73.0 & 99.2           & \textbf{0.63} \\
Guided-Retry  &                              & 35.2 & \textbf{81.0} & \textbf{99.5} & 0.71 \\
\midrule
Naive         & \multirow{3}{*}{Qwen-2.5}    & 28.7 & 62.3 & 99.5           & 0.59 \\
Inform        &                              & 21.8 & 65.5 & 99.2           & 0.45 \\
Guided-Retry  &                              & \textbf{13.2} & \textbf{76.5} & \textbf{100.0} & \textbf{0.30} \\
\bottomrule
\end{tabular}
\end{table}

\begin{table}[t]
\centering
\small
\setlength{\tabcolsep}{4pt}
\caption{SGD overall results. HR = Hallucination Rate (\%),
AAR = Appropriate Action Rate (\%),
CSR = automatic Commitment Safety Rate (\%),
UFS = User Friction Score.
Best values per model and metric are in \textbf{bold}.}
\label{tab:main_sgd}
\begin{tabular}{llcccc}
\toprule
\textbf{Strategy} & \textbf{Model} &
\textbf{HR}$\downarrow$ & \textbf{AAR}$\uparrow$ &
\textbf{CSR}$\uparrow$ & \textbf{UFS}$\downarrow$ \\
\midrule
Naive         & \multirow{3}{*}{DeepSeek-R1} & 33.2 & 47.5 & 99.8           & 0.96 \\
Inform        &                              & 15.2 & 53.0 & \textbf{100.0} & \textbf{0.41} \\
Guided-Retry  &                              & \textbf{5.2}  & \textbf{73.5} & 99.8           & 0.46 \\
\midrule
Naive         & \multirow{3}{*}{Gemma-2}     & 7.2  & 55.8 & \textbf{100.0} & 0.28 \\
Inform        &                              & \textbf{5.5}  & 61.3 & \textbf{100.0} & \textbf{0.22} \\
Guided-Retry  &                              & 5.8  & \textbf{81.0} & \textbf{100.0} & 0.44 \\
\midrule
Naive         & \multirow{3}{*}{Llama-3}     & 17.2 & 68.2 & \textbf{100.0} & 0.47 \\
Inform        &                              & 11.8 & 70.0 & 99.8           & \textbf{0.25} \\
Guided-Retry  &                              & \textbf{9.5}  & \textbf{97.8} & 99.8           & 0.26 \\
\midrule
Naive         & \multirow{3}{*}{Mistral}     & 18.5 & 60.0 & \textbf{100.0} & 0.37 \\
Inform        &                              & 11.8 & 64.5 & 99.8           & 0.24 \\
Guided-Retry  &                              & \textbf{8.8}  & \textbf{96.2} & 99.8           & \textbf{0.22} \\
\midrule
Naive         & \multirow{3}{*}{Phi-3}       & 32.2 & 62.3 & 98.8           & 0.66 \\
Inform        &                              & \textbf{25.2} & 66.5 & 98.8           & \textbf{0.52} \\
Guided-Retry  &                              & 37.0 & \textbf{76.2} & \textbf{99.8} & 0.76 \\
\midrule
Naive         & \multirow{3}{*}{Qwen-2.5}    & 16.8 & 58.2 & 99.5           & 0.51 \\
Inform        &                              & 14.5 & 60.2 & 99.2           & 0.39 \\
Guided-Retry  &                              & \textbf{6.8}  & \textbf{78.5} & \textbf{100.0} & \textbf{0.32} \\
\bottomrule
\end{tabular}
\end{table}

\subsection{Breakdown by Failure Type}

\begin{figure*}[t]
\centering
\begin{minipage}[t]{0.45\textwidth}
    \centering
    \includegraphics[width=\linewidth]{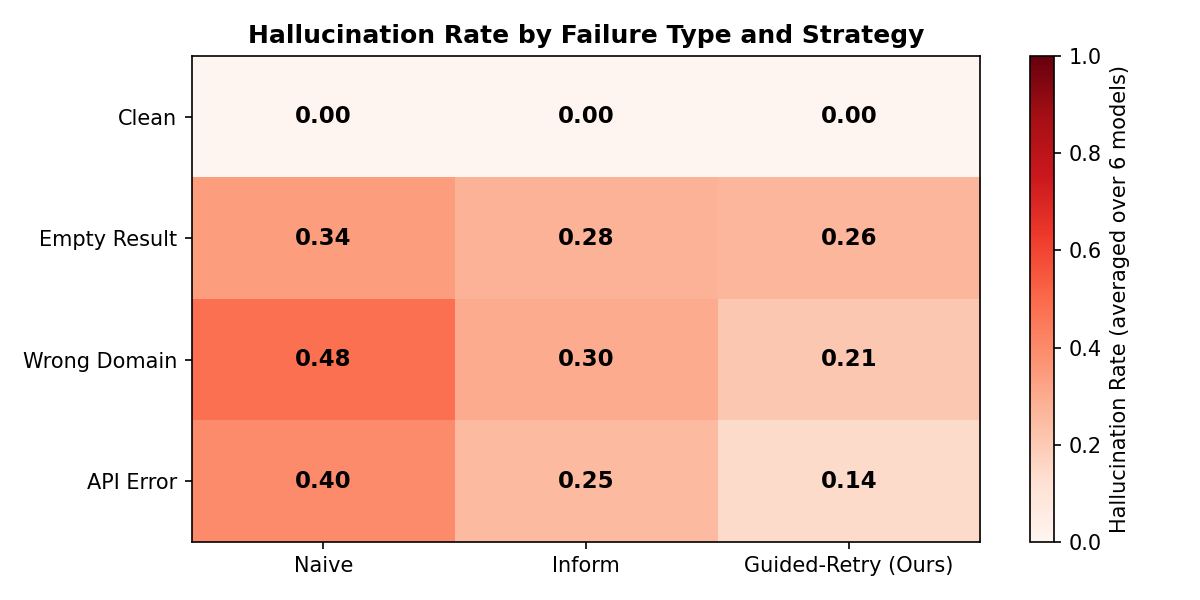}
    \vspace{-1mm}
    \small (a) MultiWOZ
\end{minipage}\hfill
\begin{minipage}[t]{0.45\textwidth}
    \centering
    \includegraphics[width=\linewidth]{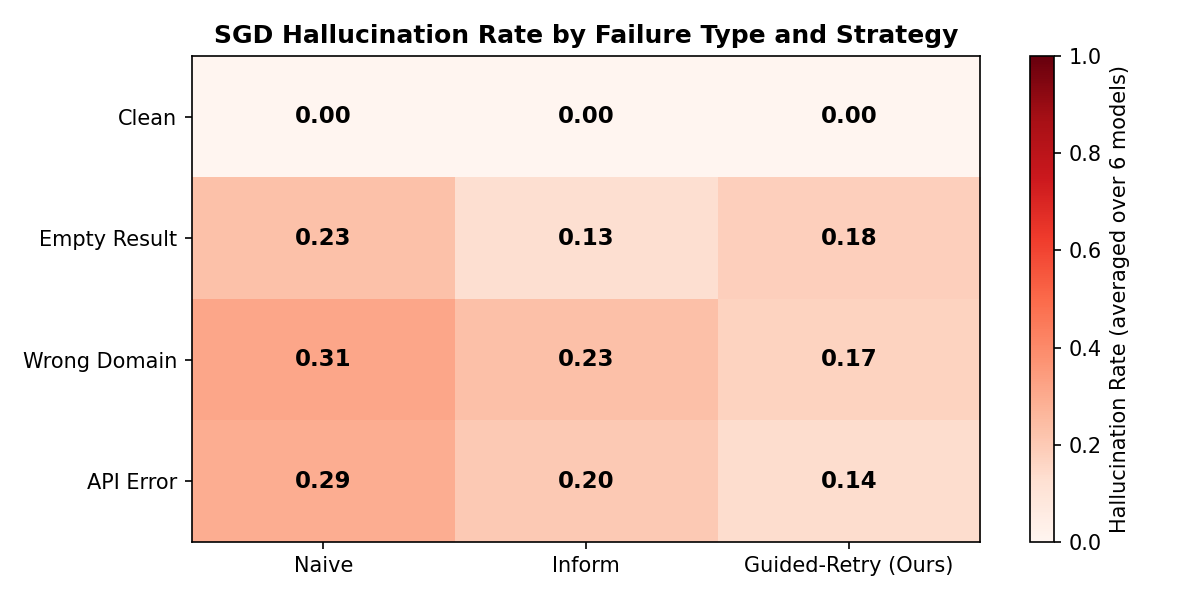}
    \vspace{-1mm}
    \small (b) SGD
\end{minipage}
\caption{Hallucination rate (\%) by failure type and recovery strategy.
Wrong-domain retrieval is the hardest failure case on both datasets, while
Guided-Retry consistently reduces hallucination relative to Naive and Inform.}
\label{fig:failure_heatmaps}
\end{figure*}

Figure~\ref{fig:failure_heatmaps} shows hallucination rates by failure type
for MultiWOZ and SGD. Wrong-domain retrieval is the hardest failure case on
both datasets. On MultiWOZ, Naive HR reaches 47.8\% for wrong-domain inputs,
and Guided-Retry still leaves 20.8\% residual hallucination. On SGD, the
same pattern holds, though at lower absolute rates: Naive HR is 31.0\% for
wrong-domain inputs, reduced to 17.0\% under Guided-Retry. This suggests
that models often adopt vocabulary from the incorrect domain result rather
than detecting the mismatch.

API errors show the clearest benefit from structured recovery prompting.
On MultiWOZ, Guided-Retry reduces HR from 39.7\% to 14.2\%; on SGD, it
reduces HR from 29.0\% to 14.0\%. Empty-result failures are also improved,
though more modestly: on MultiWOZ, HR falls from 34.3\% to 26.3\%, while on
SGD it falls from 23.0\% to 18.0\%. All strategies correctly handle clean
turns on both datasets (HR\,=\,0.0\%).

\subsection{Cross-Dataset Generalizability}

Table~\ref{tab:cross} shows results averaged across the six models on both
datasets. The overall pattern is consistent across MultiWOZ-2.2 and SGD:
Guided-Retry achieves AAR of 83.0\% and 83.9\%, respectively, confirming
that the findings generalize across structurally different TOD benchmarks
spanning 5 and 20 domains. HR is lower on SGD overall, possibly because
SGD dialogues are more formulaic, making failure status easier to detect.

\begin{table}[t]
\centering
\small
\caption{Cross-dataset comparison (6-model average).}
\label{tab:cross}
\begin{tabular}{lcccc}
\toprule
 & \multicolumn{2}{c}{\textbf{MultiWOZ-2.2}} &
   \multicolumn{2}{c}{\textbf{SGD}} \\
\cmidrule(lr){2-3}\cmidrule(lr){4-5}
\textbf{Strategy} &
\textbf{HR}$\downarrow$ & \textbf{AAR}$\uparrow$ &
\textbf{HR}$\downarrow$ & \textbf{AAR}$\uparrow$ \\
\midrule
Naive                     & 30.5 & 66.5 & 20.9 & 58.7 \\
Inform                    & 20.6 & 71.1 & 14.0 & 62.6 \\
\textbf{Guided-Retry}     & \textbf{15.3} & \textbf{83.0} &
                            \textbf{12.2} & \textbf{83.9} \\
\bottomrule
\end{tabular}
\end{table}

\subsection{Human Evaluation}

To validate the commitment-safety metric, we conducted a human evaluation
on 60 sampled responses spanning all three strategies and the three
non-clean failure types.We used nine annotators drawn from different academic levels and backgrounds to provide a diverse annotation pool, and limited the survey to 60 items to balance coverage and annotation quality. The survey took
approximately 1.5 hours per annotator to complete; a substantially longer
survey would increase annotator fatigue and could reduce labeling precision.Annotators were blind to the underlying strategy and failure condition.

Annotators received clear instructions and example items before completing
the survey. They were asked to assign a binary label (\emph{false commitment}
vs.\ \emph{appropriate response}), where a false commitment was defined as:
\emph{``the agent explicitly claims a booking, reservation, confirmation,
or specific DB-backed result as true when the DB did not return one.''}

Inter-annotator agreement was strong, with Fleiss' $\kappa = 0.7672$
and 95.74\% observed agreement. Overall human CSR was 90.0\%
(6/60 majority false commitments). The strategy ranking matched the
automatic trend: Guided-Retry achieved 100.0\% human CSR, compared with
89.47\% for Inform and 80.95\% for Naive. By failure type, human CSR was
100.0\% for empty-result cases, 85.7\% for wrong-domain cases, and 85.0\%
for API errors. These results support the validity of the automatic
commitment-safety metric while confirming that Guided-Retry yields the
safest behavior overall.

\section{Discussion}
\label{sec:discussion}

Our results show that LLM-based TOD agents can default to confident,
fluent responses even when the database explicitly reports failure,
consistent with the broader tendency of LLM agents to favor fluency over
accuracy~\citep{baidya-etal-2025-behavior}. The main practical finding is
that a single structured system-prompt addition reduces hallucination by
42--50\% without retraining, additional inference calls, or substantial
engineering overhead.

The Phi-3 result is noteworthy: it is the only model for which
Guided-Retry performs worse than Inform on HR. Phi-3 also has the highest
Naive HR (44.5\%), indicating that some models may not reliably benefit
from structured recovery instructions even when those instructions help
other model families.

Most importantly, residual hallucination remains non-trivial even under
Guided-Retry. The best model (DeepSeek-R1) still hallucinates on 5.8\% of
failure turns, while the worst case (Phi-3) reaches 35.2\%. This suggests
that prompt-level recovery is helpful, but not sufficient on its own for
robust TOD deployment.

\section{Conclusion}
\label{sec:conclusion}

We presented a controlled fault-injection study of LLM-based TOD agents
under runtime DB failures across two benchmarks and six model families.
Guided-Retry, a structured prompt-level recovery strategy, reduces
hallucination by 42--50\% without retraining, but residual hallucination
of 6--37\% remains, with wrong-domain retrieval as the hardest failure case. Code and prompts are available in our github.\footnote{\url{https://github.com/mohammad-AJP/llm-db-failure-recovery}}

\section*{Limitations}

Our fault injection is synthetically constructed, and real-world backend
failures may follow different distributions.
Automated hallucination detection relies on heuristic patterns, and human
evaluation covers only a subset of items.
We evaluate instruction-tuned models at the 7--9B scale; larger or
proprietary models may behave differently.
Finally, we do not study multi-turn recovery dynamics after the user
responds to a failure acknowledgement, which we leave for future work.

\bibliography{references_fixed}

@inproceedings{hudevcek-dusek-2023-large,
  title     = {Are Large Language Models All You Need for Task-Oriented Dialogue?},
  author    = {Hude{\v{c}}ek, Vojt{\v{e}}ch and Du{\v{s}}ek, Ond{\v{r}}ej},
  booktitle = {Proceedings of the 24th Annual Meeting of the Special Interest Group on Discourse and Dialogue},
  year      = {2023},
  pages     = {216--228},
  url       = {https://aclanthology.org/2023.sigdial-1.21},
}

@inproceedings{eric-etal-2020-multiwoz,
  title     = {{MultiWOZ} 2.1: A Consolidated Multi-Domain Dialogue Dataset},
  author    = {Eric, Mihail and Goel, Rahul and Paul, Shachi and Kumar, Adarsh and Sethi, Abhishek and Ku, Peter and Goyal, Anuj Kumar and Agarwal, Sanchit and Gao, Shuyang and Hakkani-Tur, Dilek},
  booktitle = {Proceedings of LREC 2020},
  year      = {2020},
  pages     = {422--428},
}

@inproceedings{zang-etal-2020-multiwoz,
  title     = {{MultiWOZ} 2.2: A Dialogue Dataset with Additional Annotation Corrections},
  author    = {Zang, Xiaoxue and Rastogi, Abhinav and Sunkara, Srinivas and Gupta, Raghav and Zhang, Jianguo and Chen, Jindong},
  booktitle = {Proceedings of the 2nd Workshop on NLP for Conversational AI},
  year      = {2020},
  pages     = {109--117},
  url       = {https://aclanthology.org/2020.nlp4convai-1.13},
}

@inproceedings{rastogi-etal-2020-sgd,
  title     = {Towards Scalable Multi-Domain Conversational Agents: The Schema-Guided Dialogue Dataset},
  author    = {Rastogi, Abhinav and Zang, Xiaoxue and Sunkara, Srinivas and Gupta, Raghav and Khaitan, Pranav},
  booktitle = {Proceedings of the Thirty-Fourth AAAI Conference on Artificial Intelligence},
  year      = {2020},
  pages     = {8689--8696},
}

@article{guo2025deepseekr1,
  title   = {DeepSeek-R1: Incentivizing Reasoning Capability in LLMs via Reinforcement Learning},
  author  = {Guo, Daya and Ren, Zehui and Sha, Zhangli and Fu, Zhe and Xu, Zhean and Xie, Zhenda and Zhang, Zhengyan and Hao, Zhewen and Ma, Zhicheng and Yan, Zhigang and Wu, Zhiyu and Gu, Zihui and Zhu, Zijia and Liu, Zijun and Li, Zilin and Xie, Ziwei and Song, Ziyang and Pan, Zizheng and Huang, Zhen and Xu, Zhipeng and Zhang, Zhongyu and Zhang, Zhen},
  journal = {arXiv preprint arXiv:2501.12948},
  year    = {2025},
  url     = {https://arxiv.org/abs/2501.12948},
}

@article{gemma-team-2024-gemma2,
  title   = {Gemma 2: Improving Open Language Models at a Practical Size},
  author  = {{Gemma Team, Google DeepMind}},
  journal = {arXiv preprint arXiv:2408.00118},
  year    = {2024},
  url     = {https://arxiv.org/abs/2408.00118},
}

@article{grattafiori-etal-2024-llama3,
  title   = {The Llama 3 Herd of Models},
  author  = {{Meta Llama Team}},
  journal = {arXiv preprint arXiv:2407.21783},
  year    = {2024},
  url     = {https://arxiv.org/abs/2407.21783},
}

@article{jiang-etal-2023-mistral,
  title   = {Mistral 7B},
  author  = {Jiang, Albert Q. and Sablayrolles, Alexandre and Mensch, Arthur and Bamford, Chris and Chaplot, Devendra Singh and de las Casas, Diego and Bressand, Florian and Lengyel, Gianna and Lample, Guillaume and Saulnier, Lucile and Renard Lavaud, L\'elio and Lachaux, Marie-Anne and Stock, Pierre and Le Scao, Teven and Lavril, Thibaut and Wang, Thomas and Lacroix, Timoth\'ee and El Sayed, William},
  journal = {arXiv preprint arXiv:2310.06825},
  year    = {2023},
  url     = {https://arxiv.org/abs/2310.06825},
}

@article{abdin-etal-2024-phi3,
  title   = {Phi-3 Technical Report: A Highly Capable Language Model Locally on Your Phone},
  author  = {Abdin, Marah and Aneja, Jyoti and Awadalla, Hany and Awadallah, Ahmed and Awan, Ammar Ahmad and Bach, Nguyen and Bahree, Amit and Bakhtiari, Arash and Bao, Jianmin and Behl, Harkirat and Benhaim, Alon and Bilenko, Misha and Bjorck, Johan and Bubeck, S{\'e}bastien and Cai, Martin and Cai, Qin and Chaudhary, Vishrav and Chen, Dong and Chen, Dongdong and Chen, Weizhu and Chen, Yen-Chun and Chen, Yi-Ling and Cheng, Hao and Chopra, Parul and Dai, Xiyang and Dixon, Matthew and Eldan, Ronen and Fragoso, Victor and Gao, Jianfeng and Gao, Mei and Gao, Min and Garg, Amit and Del Giorno, Allie and Goswami, Abhishek and Gunasekar, Suriya and Haider, Emman and Hao, Junheng and Hewett, Russell J. and Hu, Wenxiang and Huynh, Jamie and Iter, Dan and Jacobs, Sam Ade and Javaheripi, Mojan and Jin, Xin and Karampatziakis, Nikos and Kauffmann, Piero and Khademi, Mahoud and Kim, Dongwoo and Kim, Young Jin and Kurilenko, Lev and Lee, James R. and Lee, Yin Tat and Li, Yuanzhi and Li, Yunsheng and Liang, Chen and Liden, Lars and Lin, Xihui and Lin, Zeqi and Liu, Ce and Liu, Liyuan and Liu, Mengchen and Liu, Weishung and Liu, Xiaodong and Luo, Chong and Madan, Piyush and Mahmoudzadeh, Ali and Majercak, David and Mazzola, Matt and Mendes, Caio C{\'e}sar Teodoro and Mitra, Arindam and Modi, Hardik and Nguyen, Anh and Norick, Brandon and Patra, Barun and Perez-Becker, Daniel and Portet, Thomas and Pryzant, Reid and Qin, Heyang and Radmilac, Marko and Ren, Liliang and de Rosa, Gustavo and Rosset, Corby and Roy, Sambudha and Ruwase, Olatunji and Saarikivi, Olli and Saied, Amin and Salim, Adil and Santacroce, Michael and Shah, Shital and Shang, Ning and Sharma, Hiteshi and Shen, Yelong and Shukla, Swadheen and Song, Xia and Tanaka, Masahiro and Tupini, Andrea and Vaddamanu, Praneetha and Wang, Chunyu and Wang, Guanhua and Wang, Lijuan and Wang, Shuohang and Wang, Xin and Wang, Yu and Ward, Rachel and Wen, Wen and Witte, Philipp and Wu, Haiping and Wu, Xiaoxia and Wyatt, Michael and Xiao, Bin and Xu, Can and Xu, Jiahang and Xu, Weijian and Xue, Jilong and Yadav, Sonali and Yang, Fan and Yang, Jianwei and Yang, Yifan and Yang, Ziyi and Yu, Donghan and Yuan, Lu and Zhang, Chenruidong and Zhang, Cyril and Zhang, Jianwen and Zhang, Li Lyna and Zhang, Yi and Zhang, Yue and Zhang, Yunan and Zhou, Xiren},
  journal = {arXiv preprint arXiv:2404.14219},
  year    = {2024},
  url     = {https://arxiv.org/abs/2404.14219},
}

@article{yang-etal-2024-qwen25,
  title   = {Qwen2.5 Technical Report},
  author  = {{Qwen Team}},
  journal = {arXiv preprint arXiv:2412.15115},
  year    = {2024},
  url     = {https://arxiv.org/abs/2412.15115},
}

@article{hou-etal-2025-trace,
  title   = {Multi-Faceted Evaluation of Tool-Augmented Dialogue Systems},
  author  = {Hou, Zhaoyi Joey and Shourya, Tanya and Wang, Yingfan and Roy, Shamik and Kumar, Vinayshekhar Bannihatti and Gangadharaiah, Rashmi},
  journal = {arXiv preprint arXiv:2510.19186},
  year    = {2025},
}

@article{gupta-2026-reliability,
  title   = {{ReliabilityBench}: Evaluating {LLM} Agent Reliability Under Production-Like Stress Conditions},
  author  = {Gupta, Aayush},
  journal = {arXiv preprint arXiv:2601.06112},
  year    = {2026},
}

@article{vuddanti-etal-2025-paladin,
  title   = {{PALADIN}: Self-Correcting Language Model Agents to Cure Tool-Failure Cases},
  author  = {Vuddanti, Sri Vatsa and Shah, Aarav and Chittiprolu, Satwik Kumar and Song, Tony and Dev, Sunishchal and Zhu, Kevin and Chaudhary, Maheep},
  journal = {arXiv preprint arXiv:2509.25238},
  year    = {2025},
}

@article{shim-etal-2025-noncollaborative,
  title   = {Non-Collaborative User Simulators for Tool Agents},
  author  = {Shim, Jeonghoon and Song, Woojung and Jin, Cheyon and Kook, Seungwon and Jo, Yohan},
  journal = {arXiv preprint arXiv:2509.23124},
  year    = {2025},
}

@inproceedings{baidya-etal-2025-behavior,
  title     = {The Behavior Gap: Evaluating Zero-shot {LLM} Agents in Complex Task-Oriented Dialogs},
  author    = {Baidya, Avinash and Das, Kamalika and Gao, Xiang},
  booktitle = {Findings of ACL 2025},
  year      = {2025},
}

\end{document}